\documentclass[11pt,a4paper]{article}
\usepackage[hyperref]{acl2019}
\usepackage{times}
\usepackage{latexsym}

\usepackage{url}

\usepackage{microtype}
\usepackage{graphicx}
\usepackage{booktabs} 
\usepackage{multirow}
\usepackage{amsmath,amsthm,amssymb,amsfonts,bbm,stmaryrd}
\usepackage{bm}
\DeclareMathOperator*{\argmax}{argmax}
\usepackage{caption}
\usepackage{subcaption}
\usepackage[linguistics]{forest}

\newcommand{\bw}{\bm{w}}
\newcommand{\ba}{\bm{a}}
\newcommand{\bb}{\bm{b}}
\newcommand{\bc}{\bm{c}}
\newcommand{\br}{\bm{r}}
\newcommand{\bi}{\bm{i}}
\newcommand{\bo}{\bm{o}}
\newcommand{\bg}{\bm{g}}
\newcommand{\bW}{\bm{W}}
\newcommand{\bh}{\bm{h}}
\newcommand{\bx}{\bm{x}}

\newcommand{\bmu}{\bm{\mu}}
\newcommand{\bsi}{\bm{\Sigma}}

\newcommand{\ins}{s_{\textbf{I}}}
\newcommand{\outs}{s_{\textbf{O}}}
\newcommand{\Ra}{\shortrightarrow}
\newcommand{\Cl}{\mathcal{L}}
\newcommand{\Cn}{\mathcal{N}}
\newcommand{\iprod}[1]{\left\langle{#1}\right\rangle}

\aclfinalcopy 
\def\aclpaperid{2683} 

\title{Latent Variable Sentiment Grammar\thanks{ \,\, Work was done when the first author was visiting Westlake University. The third author is the corresponding author.}}

\author{Liwen Zhang$^\dagger$, Kewei Tu$^\dagger$, Yue Zhang$^{\ddagger \diamond}$ \\
	$^\dagger$School of Information Science and Technology,ShanghaiTech University, Shanghai, China \\
	$^\ddagger$Institute of Advanced Technology, Westlake Institute for Advanced Study, China\\
	$^\diamond$School of Engineering, Westlake University, Hangzhou, China \\
	{\tt \{zhanglw1,tukw\}@shanghaitech.edu.cn} \\
	{\tt yuezhang@westlake.edu.cn} \\
	}

\date{}

\begin{document}
\maketitle
\begin{abstract}
Neural models have been investigated for sentiment classification over constituent trees. 
They learn phrase composition automatically by encoding tree structures but do not explicitly model sentiment composition, which requires to encode sentiment class labels. 
To this end, we investigate two formalisms with deep sentiment representations that capture sentiment subtype expressions by latent variables and Gaussian mixture vectors, respectively. 
Experiments on Stanford Sentiment Treebank (SST) show the effectiveness of sentiment grammar over vanilla neural encoders. 
Using ELMo embeddings, our method gives the best results on this benchmark.
\end{abstract}

\section{Introduction}
Determining the sentiment polarity at or below the sentence level is an important task in natural language processing.
Sequence structured models \cite{li2015tree, mccann2017learned} have been exploited for modeling each phrase independently.
Recently, tree structured models \cite{zhu2015long, tai2015improved, teng2017head} were leveraged for learning phrase compositions in sentence representation given the syntactic structure.
Such models classify the sentiment over each constituent node according to its hidden vector through tree structure encoding.

\begin{figure} 
	\begin{minipage}[ht]{\linewidth} 
		\centering 
		\includegraphics[width=\linewidth]{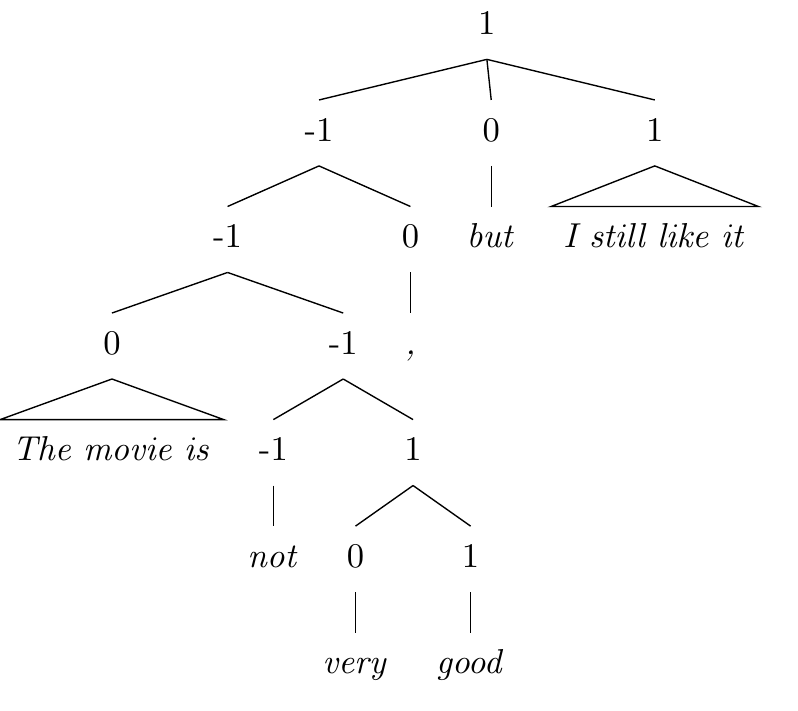}
		\caption{Example of sentiment composition} 
		\label{fig:tree_examples} 
	\end{minipage} 
\end{figure}


Though effective, existing neural methods do not consider explicit \emph{sentiment compositionality}  \cite{Montague1974-MONFPS}.
Take the sentence ``{The movie is not very good, but I still like it}'' in Figure~\ref{fig:tree_examples} as example \cite{dong2015statistical}, over the constituent tree, sentiment signals can be propagated from leaf nodes to the root, going through negation, intensification and contrast  according to the context. 
Modeling such signal channels can intuitively lead to more interpretable and reliable results.
To model sentiment composition, direct encoding of sentiment signals (e.g., +1/-1 or more fine-grained forms) is necessary.

To this end, we consider a neural network grammar with latent variables. 
In particular, we employ a grammar 
as the backbone of our approach in which nonterminals represent sentiment signals and grammar rules specify sentiment compositions. 
In the simplest version of our approach, nonterminals are sentiment labels from SST directly, resulting in a weighted grammar. 
To model more fine-grained emotions \cite{Ortony1990-ORTWBA}, we consider a latent variable grammar (LVG, \citet{matsuzaki2005probabilistic}, \citet{petrov2006learning}), which splits each nonterminal into subtypes to represent subtle sentiment signals and uses a discrete latent variable to denote the sentiment subtype of a phrase.
Finally, inspired by the fact that sentiment can be modeled with a low dimensional continuous space \cite{mehrabian1980basic}, 
we introduce a Gaussian mixture latent vector grammar (GM-LVeG, \citet{zhao2018gaussian}), which associates each sentiment signal with a continuous vector instead of a discrete variable.

Experiments on SST show that explicit modeling of sentiment composition leads to significantly improved performance over standard tree encoding, 
and models that learn subtle emotions as hidden variables give better results than coarse-grained models. 
Using a bi-attentive classification network \cite{peters2018deep} as  the encoder, out final model gives the best results on SST. 
To our knowledge, we are the first to consider neural network grammars with latent variables for sentiment composition. 
Our code will be released at \href{https://github.com/Ehaschia/bi-tree-lstm-crf}{https://github.com/Ehaschia/bi-tree-lstm-crf}. 

\section{Related Work} 
\paragraph{Phrase-level sentiment analysis}
\citet{li2015tree} and \citet{mccann2017learned} proposed sequence structured models 
that predict the sentiment polarities of the individual phrases in a sentence independently.
\citet{zhu2015long}, \citet{le2015compositional}, \citet{tai2015improved} and \citet{gupta2018attend} proposed Tree-LSTM models to capture bottom-up dependencies between constituents for sentiment analysis.
In order to support information flow bidirectionally over trees, \citet{teng2017head} introduced a Bi-directional Tree-LSTM model that adds a top-down component after Tree-LSTM encoding.
These models handle sentiment composition implicitly and predict sentiment polarities only based on embeddings of current nodes.
In contrast, we model sentiment explicitly.

\paragraph{Sentiment composition}
\citet{moilanen2007sentiment} introduced a seminal model for sentiment composition \cite{Montague1974-MONFPS}, composed positive, negative and neutral (+1/-1/0) singles hierarchically. 
\citet{taboada2011lexicon} proposed a lexicon-based method for addressing sentence level contextual valence shifting phenomena such as negation and intensification.
\citet{choi2008learning} used a structured linear model to learn semantic compositionality relying on a set of manual features.
\citet{dong2015statistical} developed a statistical parser to learn the sentiment structure of a sentence.
Our method is similar in that grammars are used to model semantic compositionality. 
But we consider neural methods instead of statistical methods for sentiment composition.
\citet{teng2016context} proposed a simple weighted-sum model of introducing sentiment lexicon features to LSTM for sentiment analysis. They used -2 to 2 represent sentiment polarities.
In contrast, we model sentiment subtypes with latent variables and combine the strength of neural encoder and hierarchical sentiment composition.

\paragraph{Latent Variable Grammar}
There has been a line of work using discrete latent variables to enrich coarse-grained constituent labels in phrase-structure parsing \cite{johnson1998pcfg, matsuzaki2005probabilistic, petrov2006learning, petrov2007improved}. 
Our work is similar in that discrete latent variables are used to model sentiment polarities. 
To our knowledge, we are the first to consider modeling fine-grained sentiment signals by investigating different types of latent variables.
Recently, there has been work using continuous latent vectors for modeling syntactic categories \cite{zhao2018gaussian}. 
We consider their grammar also in modeling sentiment polarities.

\section{Baseline}

We take the constituent Tree-LSTM as our baseline, which extends sequential LSTM to tree-structured network topologies. 
Formally, our model computes a parent representation from its two children in a Tree-LSTM:
\begin{eqnarray}
\begin{bmatrix}
\bi \\ 
\bm{f}_l \\ 
\bm{f}_r \\
\bo \\
\bg \\
\end{bmatrix} 
= 
\begin{bmatrix} 
\sigma \\
\sigma \\
\sigma \\
\sigma \\
\tanh
\end{bmatrix} 
\begin{pmatrix}
\bW_{t} 
\begin{bmatrix}
\bx \\
\bh_l \\
\bh_r 
\end{bmatrix}
 +  \bb_{t}
\end{pmatrix}
\end{eqnarray}
\begin{eqnarray}
\bc_p = \bi \otimes \bg + \bm{f}_l \otimes \bc_l + \bm{f}_r \otimes \bc_r
\end{eqnarray}

\begin{eqnarray}
\bh_p = \bo \otimes \tanh (\bc_p)
\label{eq:h_vector}
\end{eqnarray}
where $\bW_{t} \in \mathbb{R}^{5D_h \times 3D_h}$ and $\bb_{t} \in \mathbb{R}^{3D_h}$ are trainable parameters, $\otimes$ is the Hadamard product and $\bx$ represents the input of leaf node. Our formulation is a special case of the $N$-ary Tree-LSTM \cite{tai2015improved} with $N=2$.

Existing work (\citet{tai2015improved}, \citet{zhu2015long}) performs softmax classification on each node according to the state vetcor $\bh$ on each node for sentiment analysis. We follow this method in our baseline model.
\section{Sentiment Grammars}
We investigate sentiment grammars as a layer of structured representation on top of a tree-LSTM, which model the correlation between different sentiment labels over a tree. 
Depending on how a sentiment label is represented, we develop three increasingly complex models. 
In particular, the first model, which is introduced in Section~\ref{sec:wg}, uses a weighted grammar to model the first-order correlation between sentiment labels in a tree. 
It can be regarded as a PCFG model. 
The second model, which is introduced in Section~\ref{sec:lvg}, introduces a discrete latent variable for a refined representation of sentiment classes. 
Finally, the third model, which is introduced in Section~\ref{sec:lveg}, considers a continuous latent representation of sentiment classes.

\subsection{Weighted Grammars}\label{sec:wg}
\begin{figure}[!t] 
		\centering 
		\includegraphics[width=\linewidth]{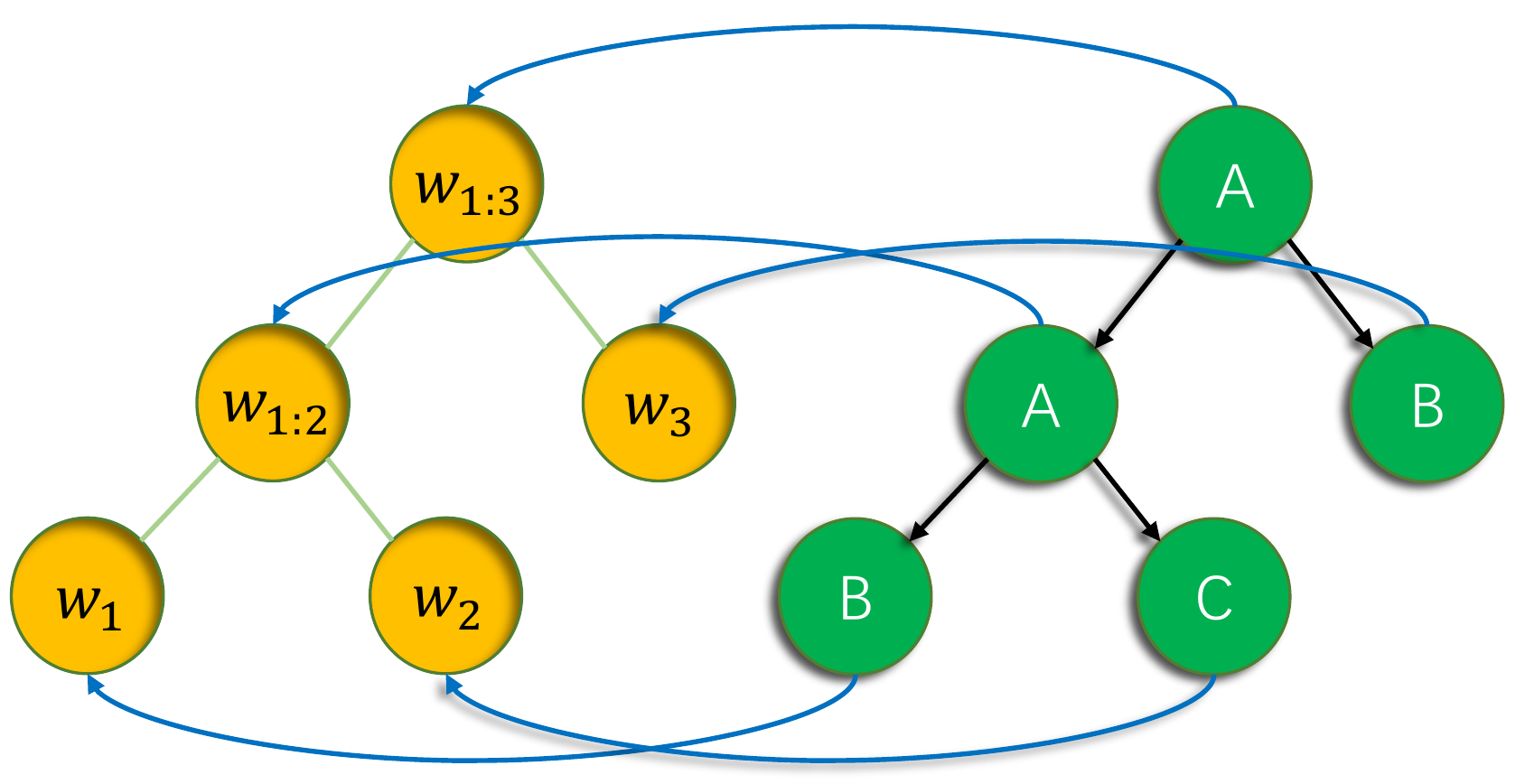}
		\caption{Sentiment grammar example. Here yellow nodes are leaf nodes of green constituent nodes, blue line $B\Ra w_1$ and black line $A \Ra BC$ represent an emission rule and a transition rule, respectively.}
		\label{fig:grammar} 
\end{figure}

Formally, a sentiment grammar is defined as $\mathcal{G} = (N, S, \Sigma, R_t, R_e, W_t, W_e)$, 
where $N = \{A, B, C, ... \}$ is a finite set of sentiment polarities, 
$S\in N$ is the start symbol, 
$\Sigma$ is a finite set of terminal symbols representing words such that $N \cap \Sigma = \varnothing$, 
$R_t$ is the transition rule set containing production rules of the form $X\Ra \alpha $ where $X \in N$ and $ \alpha \in N^+$; 
$R_e$ is the emission rule set containing production rules of the form $X \Ra \bw $ where $X \in N$ and $\bw \in \Sigma^+$. 
$W_t$ and $W_e$ are sets of weights indexed by production rules in $R_t$ and $R_e$,  respectively. 
Different from standard formal grammars, for each sentiment polarity in a parse tree our sentiment grammar invokes one emission rule to generate a string of terminals 
and invokes zero or one transition rule to product its child sentiment polarities. 
This is similar to the behavior of hidden Markov models. 
Therefore, in a parse tree each non-leaf node is a sentiment polarity and is connected to exactly one leaf node which is a string of terminals.
The terminals that are connected to the parent node can be obtained by concatenating the leaf nodes of its child nodes.
Figure~\ref{fig:grammar} shows an example for our sentiment grammar.
In this paper, we only consider $R_t$ in the Chomsky normal form (CNF) for clarity of presentation.
However, it is straightforward to extend our formulation to the general case.

The score of a sentiment tree $T$ conditioned on a sentence $\bw$ is defined as follows:
\begin{eqnarray}\label{eq:wcfg_tree}
S(T|\bw, K) = \underset{r_t \in T}{\prod} W_n(r_t) \times \underset{r_e \in T}{\prod} W_e(r_e)
\end{eqnarray}
where $r_t$ and $r_e$ represent a transition rule and an emission rule in sentiment parse tree $T$, respectively.
We specify the transition weights $W_n$ with a non-negative rank-3 tensor.
We compute the non-negative weight of each emission rule $W_e(X \Ra \bw_{i:j})$ by applying a single layer perceptron $f_X$ and an exponential function to the neural encoder state vector $\bh_{i:j}$ representing the constituent $\bw_{i:j}$.

Sentiment grammars provides a principled way for explicitly modeling sentiment composition, and through parameterizing the emission rules with neural encoders, it can take the advantage of deep learning.
In particular, by adding a weighted grammar on top of a tree-LSTM, our model is reminiscent of LSTM-CRF in the sequence structure. 

\subsection{Latent Variable Grammars}\label{sec:lvg}
Inspired by categorical models \cite{Ortony1990-ORTWBA} which regard emotions as an overlay over a series of basic emotions,
we extend our sentiment grammars with Latent Variable Grammars (LVGs; \citet{petrov2006learning}), which refine each constituent tree node with a discrete latent variables, splitting each observed sentiment polarity into finite unobserved sentiment subtypes. We refer to trees over unsplit sentiment polarities as \emph{unrefined trees} and trees over sentiment subtypes as \emph{refined trees}.

Suppose that the sentiment polarities $A$, $B$ and $C$ of a transition rule $A \Ra BC$ are split into $n_A$, $n_B$ and $n_C$ subtypes, respectively. 
The weights of the refined transition rule can be represented by a non-negative rank-3 tensor $W_{A \Ra BC} \in \mathbb{R}^{n_A \times n_B \times n_C}$.
Similarly, given an emission rule $A \Ra \bw_{i:j}$, the weights of its refined rules by splitting $A$ into $n_A$ subtypes is a non-negative vector $W_{A \Ra \bw_{i:j}} \in \mathbb{R}^{n_A}$ calculated by  an exponential function and a single layer perceptron $f_A$: 
\begin{eqnarray}
W_{A \Ra \bw_{i:j}} = \exp(f_A(\bh_{i:j}))
\end{eqnarray}
where $\bh_{i:j}$ is the vector representation of constituent $\bw_{i:j}$.
The score of a refined parse tree is defined as the product of weights of all transition rules and emission rules that make up the refined parse tree, similar to Equation~\ref{eq:wcfg_tree}.
The score of an unrefined parse tree is then defined as the sum of the scores of all refined trees that are consistent with it.

Note that Weighted Grammar (WG) can be viewed as a special case of LVGs where each sentiment polarity has one subtype.

\subsection{Gaussian Mixture Latent Vector Grammars}\label{sec:lveg}
Inspired by continuous models \cite{mehrabian1980basic} which model emotions in a continuous low dimensional space,
we employ Latent Vector Grammars (LVeGs) \cite{zhao2018gaussian} that associate each sentiment polarity with a latent vector space representing the set of sentiment subtypes. 
We follow the idea of Gaussian Mixture LVeGs (GM-LVeGs) \cite{zhao2018gaussian}, which uses Gaussian mixtures to model weight functions. 
Because Gaussian mixtures have the nice property of being closed under product, summation, and marginalization, 
learning and parsing can be done efficiently using dynamic programming

In GM-LVeG, the weight function of a transition or emission rule $r$ is defined as a Gaussian mixture with $K_r$ mixture components:
\begin{eqnarray}
W_{r}(\br) = \sum_{k=1}^{K_{r}} \rho_{r, k} \Cn(\br | \bmu_{r, k}, \bsi_{r, k})
\end{eqnarray}
where $\br$ is the concatenation of the latent vectors representing subtypes for sentiment polarities in rule $r$, 
 $\rho_{r, k} > 0 $ is the $k$-th mixing weight (the $K_r$ mixture weights do not necessarily sum up to 1),  
and $\Cn(\br | \bmu_{r, k}, \bsi_{r, k})$ denotes the $k$-th Gaussian distribution parameterized by mean $\bmu_{r, k}$ and co-variance matrix $\bsi_{r, k}$.
For an emission rule $A \Ra \bw_{i:j}$, all the Gaussian mixture parameters are calculated by single layer perceptrons from the vector representation $\bh_{i:j}$ of constituent $\bw_{i:j}$: 
\begin{eqnarray}
\rho_{r, k} & = & \exp(f_A^{\rho, k}(\bh_{i:j})) \nonumber \\
\bmu_{r, k} & = & f_A^{\bmu, k}(\bh_{i:j}) \\
\bsi_{r, k} & = & \exp(f_A^{\bsi, k}(\bh_{i:j})) \nonumber
\end{eqnarray}

For the sake of computational efficiency, we use Gaussian distributions with diagonal co-variance matrices. 

\subsection{Parsing} \label{sec:parsing}
The goal of our task is to find the most probable sentiment parse tree $T^*$, 
given a sentence $\bw$ and its constituency parse tree skeleton $K$. 
The polarity of the root node represents the polarity of the whole sentence, 
and the polarity of a constituent node is considered as the polarity of the phrase spanned by the node. 
Formally, $T^*$ is defined as:
\begin{eqnarray}\label{eq:parse_obj}
    T^* = \argmax_{T \in G(\bw, K)} P(T | \bw, K)
\end{eqnarray}
where $G(\bw, K)$ denotes the set of unrefined sentiment parse trees for $\bw$ with skeleton $K$. 
$P(T | w, K)$ is defined based on the parse tree score Equation~\ref{eq:wcfg_tree}: 
\begin{eqnarray}
P(T|\bw, K) = \frac{S(T| \bw, K)}{\sum_{\hat{T} \in K} S(\hat{T} | \bw, K)} \,.
\end{eqnarray}
Note that unlike syntactic parsing, on SST we do not need to consider structural ambiguity, and thus resolving only rule ambiguity. 

$T^*$ can be found using dynamic programming such as the CYK algorithm for WG.
However, parsing becomes intractable with LVGs and LVeGs since we have to compute the score of an unrefined parse tree by summing over all of its refined versions. 
We use the best performing max-rule-product decoding algorithm \cite{petrov2006learning, petrov2007improved} for approximate parsing, 
which searches for the parse tree that maximizes the product of the posteriors (or expected counts) of unrefined rules in the parse tree.
The detailed procedure is described below, which is based on the classic inside-outside algorithm.

\begin{table*}[!ht]
	\centering
	{\setlength{\tabcolsep}{.0em}
		\begin{tabular}{c}
			\toprule 
			\begin{minipage}{\linewidth}
				\vspace{-.5em}
				\begin{eqnarray}
				\label{eq:lvg_inside}
				\ins^{A}(a, i, j) =& \underset{A\Ra BC \in R_t}{\sum} 
				& \underset{b \in N}{\sum} \,\, \underset{c \in N}{\sum} \,\, W_{A\Ra \bw_{i:j}}(a) \,\, W_{A\Ra BC}(a,b,c) \times \ins^{B}(b, i,k) \,\, \ins^{C}(c, k + 1,j) \,. \\ 
				\label{eq:lvg_outside} 
				\outs^{A}(a, i, j) =& \underset{B\Ra CA \in R_t}{\sum} 
				& \underset{b \in N}{\sum} \,\, \underset{c \in N}{\sum} \,\, W_{A\Ra \bw_{i:j}}(a) \,\, W_{B\Ra CA}(b,c,a) \times \outs^{B}(b, k, j) \,\, \ins^{C}(c, k, i - 1) \nonumber \\
				+ & \underset{B\Ra AC \in R_t}{\sum} & \underset{b \in N}{\sum} \,\, \underset{c \in N}{\sum} \,\, W_{A\Ra \bw_{i:j}}(a) \,\, W_{B\Ra AC}(b,a,c) \times \outs^{B}(b, i, k) \,\, \ins^{C}(c, j + 1, k) \,.
				\end{eqnarray}
				\vspace{-.5em}
			\end{minipage}\\
			\midrule
			\begin{minipage}{\linewidth}
				\vspace{-.5em}
				\begin{eqnarray}
				\label{eq:lvg_rscore}
				s(A\Ra BC, i, k, j) =\! \underset{a \in N}{\sum} \,\, \underset{b \in N}{\sum} \,\, \underset{c \in N}{\sum}  \,\, W_{A\Ra BC}(a,b,c) \times \outs^{A}(a, i, j) \times \ins^{B}(b, i, k) \times \ins^{C}(c, k+1, j) \, .
				\end{eqnarray}
				\vspace{-.5em}
			\end{minipage}\\
			\toprule
			\begin{minipage}{\linewidth}
				\vspace{-.5em}
				\begin{eqnarray}
				\label{eq:inside}
				\ins^{A}(\ba, i, j) =& \underset{A\Ra BC \in R_t}{\sum}
				& \iint W_{A\Ra \bw_{i:j}}(\ba) W_{A\Ra BC}(\ba,\bb,\bc) \times \ins^{B}(\bb, i,k) \ins^{C}(\bc, k + 1,j) \,d\bb d\bc\,. \\ 
				\label{eq:outside} 
				\outs^{A}(\ba, i, j) =& \underset{B\Ra CA \in R_t}{\sum}
				& \iint W_{A\Ra \bw_{i:j}}(\ba) W_{B\Ra CA}(\bb,\bc,\ba) \times \outs^{B}(\bb, k, j) \ins^{C}(\bc, k, i - 1) \,d\bb d\bc \nonumber \\
				+& \underset{B\Ra AC \in R_t}{\sum} \,\, & \iint W_{A\Ra \bw_{i:j}}(\ba) W_{B\Ra AC}(\bb,\ba,\bc) \times \outs^{B}(\bb, i, k) \ins^{C}(\bc, j + 1, k) \,d\bb d\bc.
				\end{eqnarray}
				\vspace{-.5em}
			\end{minipage}\\
			\midrule
			\begin{minipage}{\linewidth}
				\vspace{-.5em}
				\begin{eqnarray}
				\label{eq:rscore}
				s(A\Ra BC, i, k , j) =\! \iiint W_{A\Ra BC}(\ba,\bb,\bc) \times \outs^{A}(\ba, i, j) \times \ins^{B}(\bb, i, k) \times \ins^{C}(\bc, k + 1, j) \,d\ba d\bb d\bc\,. \!\!\!\!
				\end{eqnarray}
				\vspace{-.5em}
			\end{minipage}\\
			\bottomrule
	\end{tabular}}
	\caption{
	    \label{tab:eqs}
	    Equation~\ref{eq:lvg_inside}-\ref{eq:lvg_rscore} calculate the inside score, outside score and production rule score for LVG, respectively. Equation~\ref{eq:inside}-\ref{eq:rscore} is used for LVeG. 
		Equation~\ref{eq:lvg_inside} and Equation~\ref{eq:inside} are
		the inside score functions of a sentiment polarity $A$ over its span $\bw_{i:j}$ in the sentence $\bw_{1:n}$.
		Equation~\ref{eq:lvg_outside} and Equation~\ref{eq:outside} are
		the outside score functions of a sentiment polarity $A$ over a span $\bw_{i:j}$ in the sentence $\bw_{1:n}$.
		Equation~\ref{eq:lvg_rscore} and Equation~\ref{eq:rscore}:  
		the production rule score function of a rule $A\Ra BC$ with sentiment polarities $A$, $B$, and $C$ spanning words $\bw_{i:j}$, $\bw_{i, k-1}$, and $\bw_{k + 1:j}$ respectively.
		Here we use lower case letters $a$, $b$, $c\dots $ represent discrete subtypes of sentiment polarities $A$, $B$, $C \dots$ in LVG and use bold lower case letters $\ba$, $\bb$, $\bc\dots $ represent continuous subtypes of sentiment polarities in LVeG. 
		Note that spans such as $\bw_{i:j}$ mentioned above are all given by the skeleton $K$ of sentence $\bw_{1:n}$.}
\end{table*}

For LVGs, we first use dynamic programming to recursively calculate the inside score function $\ins^{A}(a, i, j)$ and outside score function $\outs^{A}(a, i, j)$ for each sentiment polarity over each span $\bw_{i:j}$ consistent with skeleton $K$ using Equation~\ref{eq:lvg_inside} and Equation~\ref{eq:lvg_outside} in Table~\ref{tab:eqs}, respectively. 
Similarly for LVeGs, we recursively calculate inside score function $\ins^{A}(\ba, i, j)$ and outside score function $\outs^{A}(\ba, i, j)$ in LVeG are calculated by Equation~\ref{eq:inside} and Equation~\ref{eq:outside} in Table~\ref{tab:eqs}, 
in which we replace the sum of discrete variables in Equation~\ref{eq:lvg_inside}-\ref{eq:lvg_outside} with the integral of continuous vectors. 
Next, using Equation~\ref{eq:lvg_rscore} and Equation~\ref{eq:rscore} in Table~\ref{tab:eqs}, 
we calculate the score $s(A\Ra BC, i, k, j)$ for LVG and LVeG, respectively, 
where $\iprod{A\Ra BC, i, k, j}$ represents an anchored transition rule $A \Ra BC$ 
with $A$, $B$ and $C$ spanning phrase $\bw_{i:j}$, $\bw_{i, k-1}$ and $\bw_{k+1:j}$ (all being consistent with skeleton $K$), respectively. 
The posterior (or expected count ) of $\iprod{A\Ra BC, i, k, j}$ can be calculate as follows:
\begin{eqnarray}\label{eq:ecount}
    q(A \Ra BC, i, k, j) = \frac{s(A \Ra, BC, i, k, j)}{\ins(S, 1, n)},
\end{eqnarray}
where $\ins(S, 1, n)$ is the inside score for the start symbol $S$ over the whole sentence $\bw_{1:n}$. 
Then we can run CYK algorithm to identify the parse tree that maximizes the product of rule posteriors.
It's objective function is given by:
\begin{eqnarray}\label{eq:mparse}
T^*_{q} = \argmax_{T_q \in G(\bw, K)} \prod_{e \in T} q(e)
\end{eqnarray}
where $e$ ranges over all the transition rules in the sentiment parse tree $T$. 

Note that the equations in Table~\ref{tab:eqs} are tailored for our sentiment grammars and differ from their standard versions in two aspects. First, we take into account the additional emission rules in the inside and outside computation; second, the parse tree skeleton is assumed given and hence the split point $k$ is prefixed in all the equations.

\subsection{Learning}
Given a training dataset $D = \{T_{i}, \bw_i, K_i| i = 1 \dots m \}$ containing $m$ samples, 
where $T_{i}$ is the gold sentiment parse tree for the sentence $\bw_i$ with its corresponding gold tree skeleton $K_i$. The discriminative learning objective is to minimize the negative log conditional likelihood:
\begin{eqnarray}\label{eq:learn_obj}
\Cl(\Theta) = -\log \prod_{i = 1}^{m} P_{\Theta}(T_{i} | \bw_i, K_i)\,,
\end{eqnarray}
where $\Theta$ represents the set of trainable parameters of our models.
We optimize the objective with gradient-based methods. 
In particular, gradients are first calculated over the sentiment grammar layer, before being back-propagated to the tree-LSTM layer.
 
The gradient computation for the three models involves computing expected counts of rules, which has been described in Section~\ref{sec:parsing}. For WG and LVG, the derivative of $W_{r}$, the parameter of an unrefined production rule $r$ is:
\begin{eqnarray}
\label{eq:lvg_grad}
\frac{ \partial \Cl(\Theta)}{\partial W_r} = \sum_{i=1}^{m} (\mathbb{E}_{\Theta}[f_r(t) | T_i] - \mathbb{E}_{\Theta}[f_r(t) |\bw_i] ), \!\!
\end{eqnarray}
where $\mathbb{E}_{\Theta}[f_r(t) | T_i]$ denotes the expected count of the unrefined production rule $r$ with respect to $P_{\Theta}$ in the set of refined trees $t$, which are consistent with the observed parse tree $T$. 
Similarly, we use $\mathbb{E}_{\Theta}[f_r(t) |\bw_i]$ for the expectation over all derivations of the sentence $\bw_i$.

For LVeG, the derivative with respect to $\Theta_r$, the parameters of the weight function $W_r(\br)$ of an unrefined production rule $r$ is:
\begin{align}
\label{eq:lveg_grad}
\frac{\partial\Cl(\Theta)}{\partial\Theta_r} \! =& \sum_{i=1}^{m} \int
\left( 
\frac{\partial W_r(\br)} {\partial\Theta_r}
\right. \\
\! \times& \left.
\frac{\mathbb{E}_{\Theta}[f_r(t) | \bw_i] - \mathbb{E}_{\Theta}[f_r(t) | T_i]}{W_r(\br)} 
\right)\,d\br\,. \nonumber
\end{align}
The two expectations in Equation~\ref{eq:lvg_grad} and \ref{eq:lveg_grad} can be efficiently computed using the inside-outside algorithm in Table~\ref{tab:eqs}.
The derivative of the parameters of neural encoder can be derived from the derivative of the parameters of the emission rules.

\begin{figure*}[htbp]
\centering
\begin{minipage}[t]{0.32\textwidth}
\centering
\includegraphics[width=5.0cm]{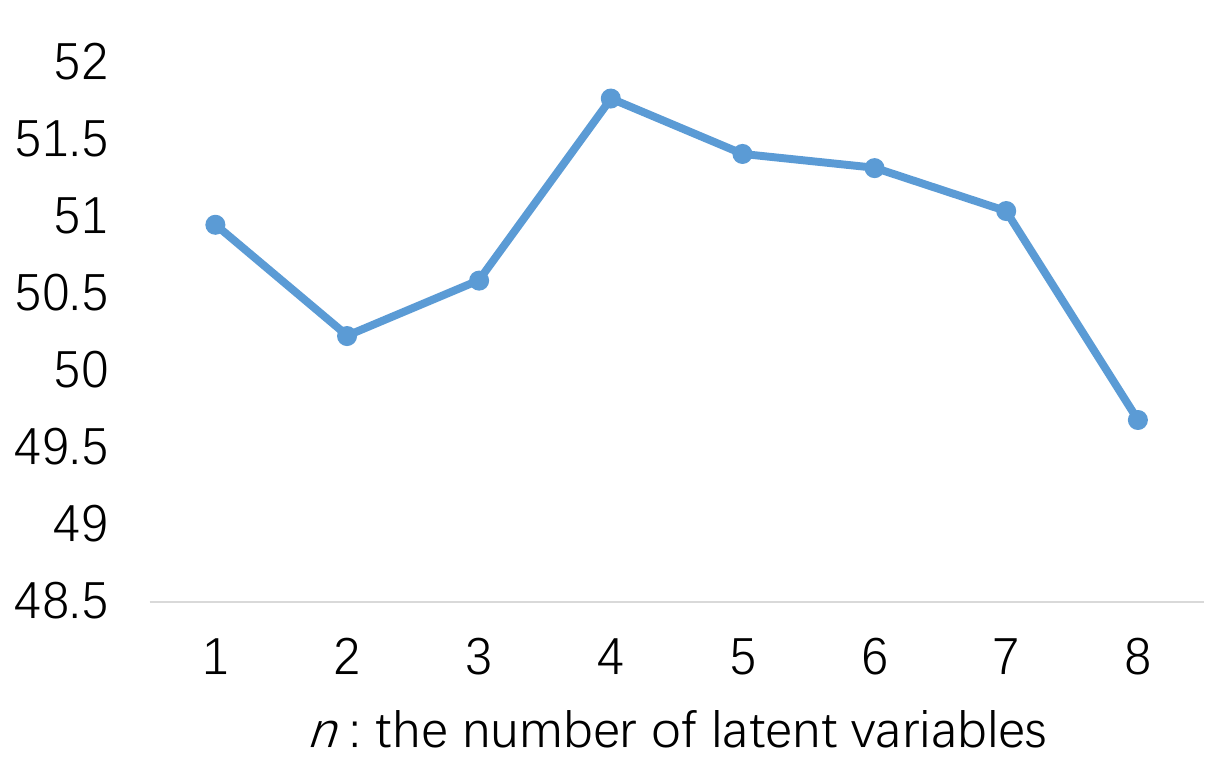}
\captionsetup{labelformat=empty}
\caption{(a)}
\end{minipage}
\begin{minipage}[t]{0.32\textwidth}
\centering
\includegraphics[width=5cm]{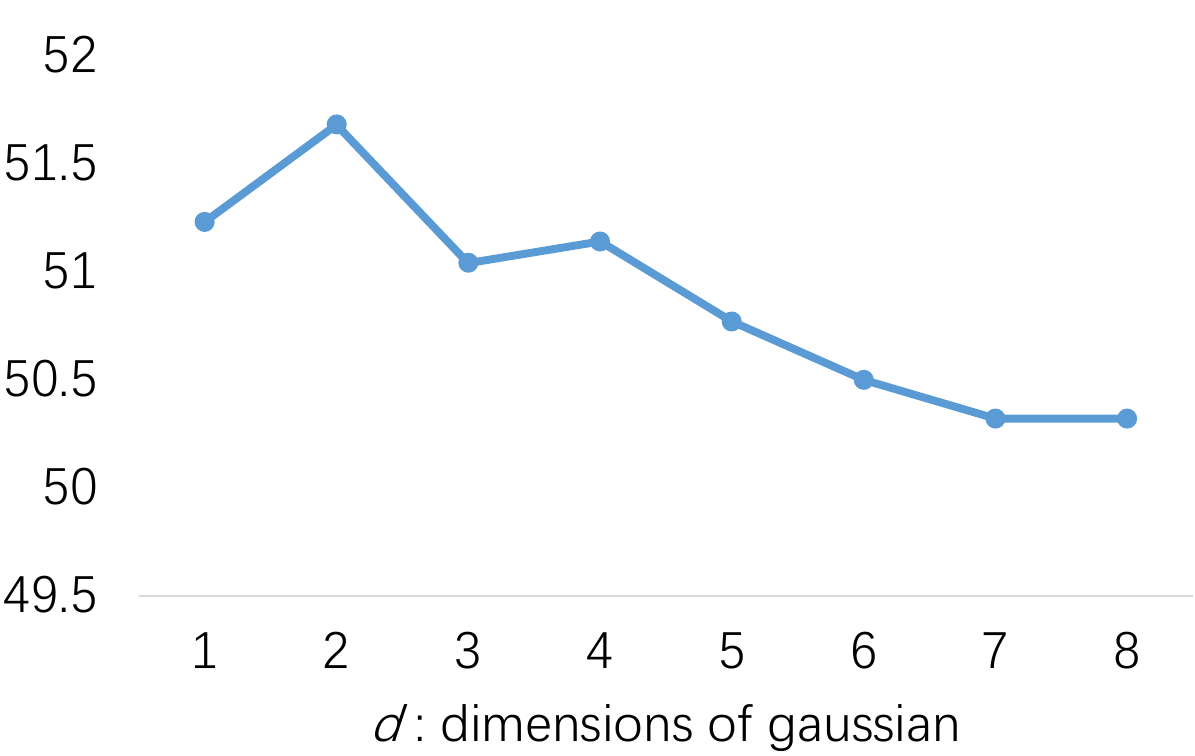}
\captionsetup{labelformat=empty}
\caption{(b)}
\end{minipage}
\begin{minipage}[t]{0.32\textwidth}
\centering
\includegraphics[width=5cm]{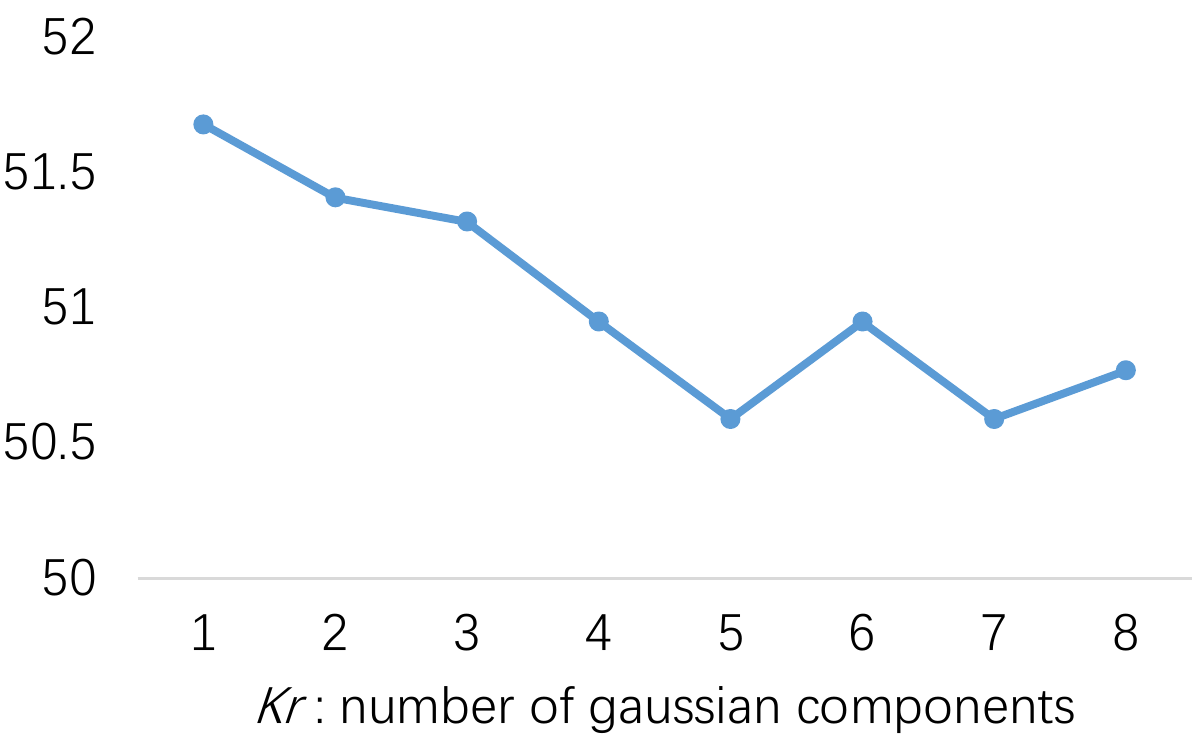}
\captionsetup{labelformat=empty}
\caption{(c)}
\end{minipage}
\caption{Sentence level accuraicies on the development dataset. Figure~(a) shows the performance of ConTree+LVG with different latent variables. Figure~(b) and Figure~(c) show the performance of ConTree+LVeG with different Gaussian dims and Gaussian mixture component numbers, respectively. }
\label{fig:dev}
\end{figure*}

\section{Experiments}
To investigate the effectiveness of modeling sentiment composition explicitly and using discrete variables or continuous vectors to model sentiment subtypes, 
we compare standard constituent Tree-LSTM (ConTree) with our models ConTree+WG, ConTree+LVG and ConTree+LVeG, respectively. To show the universality of our approaches, we also experiment with the combination of a state-of-the-art sequence structured model, bi-attentive classification network (BCN, \citet{peters2018deep}), with our model: BCN+WG, BCN+LVG and BCN+LVeG. 

\subsection{Data}
We use Stanford Sentiment TreeBank (SST, \citet{socher2013recursive}) for our experiments.
Each constituent node in a phrase-structured tree is manually assigned an integer sentiment polarity from 0 to 4, which correspond to five sentiment classes: very negative, negative, neutral, positive and very positive, respectively.
The root label represents the sentiment label of the whole sentence. 
The constituent node label represents the sentiment label of the phrase it spans. 
We perform both binary classification (-1, 1) and fine-grained classification (0-4), called SST-2 and SST-5, respectively.
Following previous work, we use the labels of all phrases and gold-standard tree structures for training and testing. 
For binary classification, we merge all positive labels and negative labels.

\subsection{Experimental Settings}
\paragraph{Hyper-parameters} For ConTree, word vectors are initialized using Glove \cite{pennington2014glove} 300-dimensional embeddings and are updated together with other parameters. 
We set the hidden size of hidden units is 300. 
Adam \cite{kingma2014adam} is used to optimize the parameters with learning rate is 0.001.
We adopt Dropout after the Embedding layer with a probability of 0.5. The sentence level mini-batch size is 32.
For BCN experiment, we follow the model setting in \citet{mccann2017learned} except the sentence level mini-batch is set to 8.

\subsection{Development Experiments}
We use the SST development dataset to investigate different configurations of our latent variables and Gaussian mixtures. The best performing parameters on the development set are used in all following experiments.

\paragraph{LVG subtype numbers}
To explore the suitable number of latent variables to model subtypes of a sentiment polarity,
we evaluate our ConTree+LVG model with different number of latent variables from 1 to 8.
Figure~\ref{fig:dev}(a) shows that there is an upward trend while the number of hidden variables $n$ increases from 1 to 4.
After reaching the peak when $n=4$, the accuracy decreases as the number of latent variable continue to increase.
We thus choose $n=4$ for remaining experiments. 

\begin{table*}[ht]
\centering
\begin{tabular}{lcccc}
\hline
              Model               & SST-5 Root & SST-5 Phrase & SST-2 Root & SST-2 Phrase  \\ \hline
ConTree \cite{le2015compositional} & 49.9 & -       & 88.0    & -   \\
ConTree \cite{tai2015improved}  & 51.0    & -       & 88.0    & -   \\
ConTree \cite{zhu2015long}      & 50.1    & -       & -       & -  \\
ConTree \cite{li2015tree}       & 50.4    & \textbf{83.4} & 86.7    & -  \\ \hline
ConTree (Our implementation)     & 51.5   & 82.8    & 89.4    & 86.9  \\ \hline
ConTree + WG                   & 51.7    & 83.0    & 89.7    & 88.9 \\
ConTree + LVG4                  & 52.2    & 83.2    & \textbf{89.8}    & 89.1 \\
ConTree + LVeG                & \textbf{52.9}  & \textbf{83.4}  & \textbf{89.8}  & \textbf{89.5}  \\ \hline
\end{tabular}
\caption{Experimental results with constituent Tree-LSTMs.}
\label{tab:main_res}
\end{table*}

\begin{table}[ht]
\centering
\begin{tabular}{lcccc}
\hline
\multirow{2}{*}{Model}    & \multicolumn{2}{c}{SST-5}       & \multicolumn{2}{c}{SST-2}       \\
                          & Root     & Phrase         & Root           & Phrase             \\ \hline
BCN(P)              & 54.7           & -              & -              & -              \\ \hline
BCN(O)              & 54.6           & 83.3           & 91.4           & 88.8           \\ \hline
BCN+WG              & 55.1           & \textbf{83.5}  & 91.5           & 90.5           \\
BCN+LVG4            & 55.5           & \textbf{83.5}  & 91.7           & 91.3  \\
BCN+LVeG            & \textbf{56.0}  & \textbf{83.5}  & \textbf{92.1}  & \textbf{91.6}  \\ \hline
\end{tabular}
\caption{Experimental results with ELMo. BCN(P) is the BCN implemented by \citet{peters2018deep}. BCN(O) is the BCN implemented by ourselves.}
\label{tab:elmo_res}
\end{table}

\paragraph{LVeG Gaussian dimensions}
We investigate the influence of the latent vector dimension on the accuracy for ConTree+LVeG.
The component number of Gaussian mixtures is fixed to 1, Figure~\ref{fig:dev}(b) illuminates that as the dimension increases from 1 to 8, there is a rise of accuracy from 1 to 2, followed by a decrease from 2 to 8. 
Thus we set the Gaussian dimension to 2 for remaining experiments.

\paragraph{LVeG Gaussian mixture component numbers}
Future~\ref{fig:dev}(c) shows the performance of different component numbers with fixing the Gaussian dimension to 2. 
With the increase of Gaussian component number, the fine-grained sentence level accuracy declines slowly.
The best performance is obtained when the component number $K_r=1$, which we choose for remaining experiments.

\subsection{Main Results}
We re-implement constituent Tree-LSTM (ConTree) of \citet{tai2015improved} and obtain better results than their original implementation.
We then integrate ConTree with Weighted Grammars (ConTree+WG), Latent Variable Grammars with a subtype number of 4 (ConTree+LVG4), and Latent Variable Grammars (ConTree+LVeG), respectively.
Table~\ref{tab:main_res} shows the experimental results for sentiment classification on both SST-5 and SST-2 at the sentence level (Root) and all nodes (Phrase).

The performance improvement of ConTree+WG over ConTree reflects the benefit of handling sentiment composition explicitly. 
Particularly the phrase level binary classification task, ConTree+WG improves the accuracy by 2 points.

Compared with ConTree+WG, ConTree+LVG4 improves the fine-grained sentence level accuracy by 0.5 point, which demonstrates the effectiveness of modeling the sentiment subtypes with discrete variables. 
Similarly, incorporating Latent Vector Grammar into the constituent Tree-LSTM, 
the performance improvements, especially on the sentence level SST-5, 
demonstrate the effectiveness of modeling sentiment subtypes with continuous vectors.
The performance improvements of ConTree+LVeG over ConTree+LVG4 show the advantage of infinite subtypes over finite subtypes. 

There has also been work using large-scale external datasets to improve performances of sentiment classification.
\citet{peters2018deep} combined bi-attentive classification network (BCN, \citet{mccann2017learned}) with a pretrained language model with character convolutions on a large-scale corpus (ELMo) and reported an accuracy of 54.7 on sentence-level SST-5. 
For fair comparison, we also augment our model with ELMo. 
Table~\ref{tab:elmo_res} shows that our methods beat the baseline on every task. 
BCN+WG improves accuracies on all task slightly by modeling sentiment composition explicitly.
The obvious promotion of BCN+LVG4 and BCN+LVeG shows that explicitly modeling sentiment composition with fine-grained sentiment subtypes is useful.
Particularly, BCN+LVeG improves the sentence level classification accurracies by 1.4 points (fine-grained) and 0.7 points (binary) compared to BCN (our implementation), respectively. 
To our knowledge, we achieve the best results on the SST dataset.

\begin{figure} 
	\begin{minipage}[ht]{\linewidth} 
		\centering 
		\includegraphics[width=\linewidth]{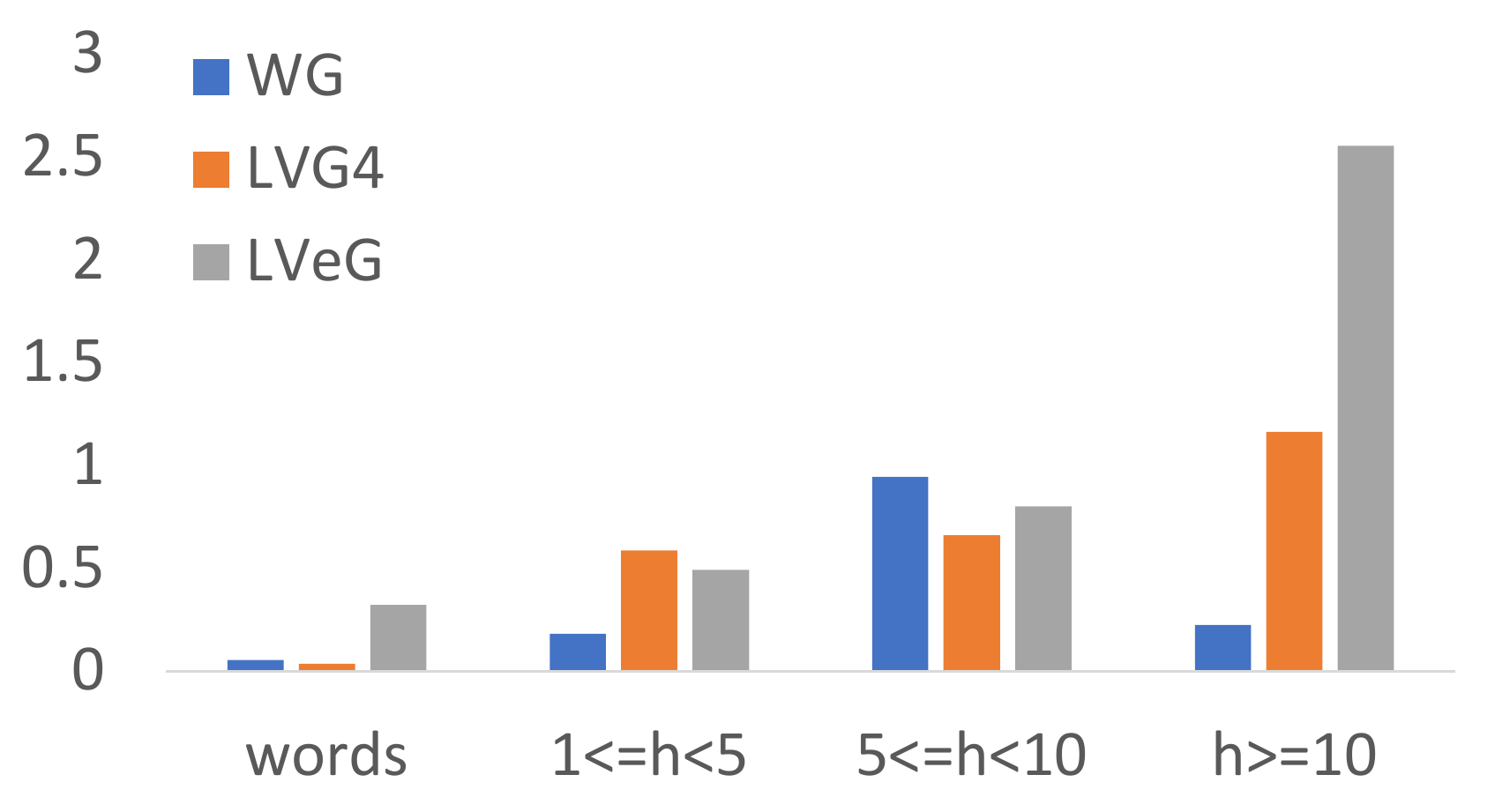}
		\caption{Changes in phrase level 5-class accuracies of our methods over ConTree.} 
		\label{fig:len_acc} 
	\end{minipage} 
\end{figure}

\subsection{Analysis}
We make further analysis of our methods based on the constituent Tree-LSTM model. In the following, using WG, LVG and LVeG denote our three methods, respectively.

\paragraph{Impact on words and phrases}
Figure~\ref{fig:len_acc} shows the accuracy improvements over ConTree on phrases of different heights.
Here the height $h$ of a phrase in parse tree is defined as the distance between its corresponding constituent node and the deepest leaf node in its subtree.
The improvement of our methods on word nodes, whose height is 0, is small because neural networks and word embeddings can already capture the emotion of words.
In fact, the accuracy of ConTree on word nodes reaches 98.1\%.
As the height increases, the performance of our methods increase, expect for the accuracies of WG when $h \geq 10$ since the coarse-grained sentiment representation is far difficulty for handling too many sentiment compositions over the tree structure. 
The performance improvements of LVG4 and LVeG when $h \geq 10$ show modeling fine-grained sentiment signals can represent sentiment of higher phrases better. 

\begin{figure} 
	\begin{minipage}[ht]{\linewidth} 
		\centering 
		\includegraphics[width=\linewidth]{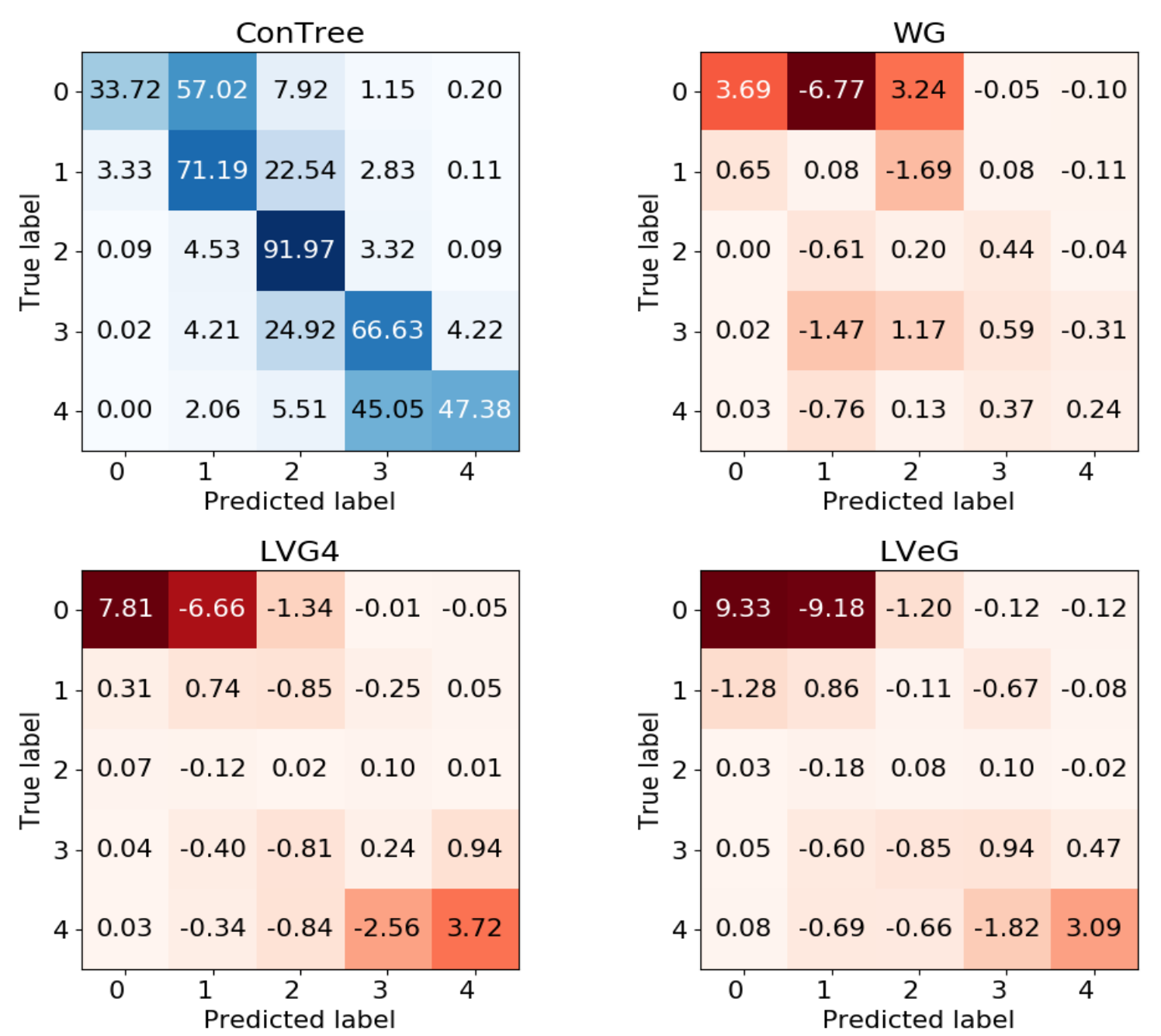}
		\caption{Top left is the normalized confusion matrix on the 5-class phrase level test dataset for ConTree. The others are the performance changs of our model over ConTree. Value in each cell is written with a unit of $\times 10^{-2}$} 
		\label{fig:confusion} 
	\end{minipage} 
\end{figure}

\paragraph{Impact on sentiment polarities}
Figure~\ref{fig:confusion} shows the performance changes of our models over ConTree on different sentiment polarities. 
The accuracy of every sentiment polarity on WG over ConTree improves slightly.
Compared with ConTree, the accuracies of LVG4 and LVeG on extreme sentiments (the strong negative and strong positive sentiments) receive significant improvement. 
In addition, the proportion of extreme emotions mis-classified as weak emotions (the negative and positive sentiments) drops dramatically. 
It indicates that LVG4 and LVeG can capture the subtle difference between extreme sentiments and weak sentiments by modeling sentiment subtypes explicitly. 

\begin{figure} 
	\begin{minipage}[ht]{\linewidth} 
		\centering 
		\includegraphics[width=\linewidth]{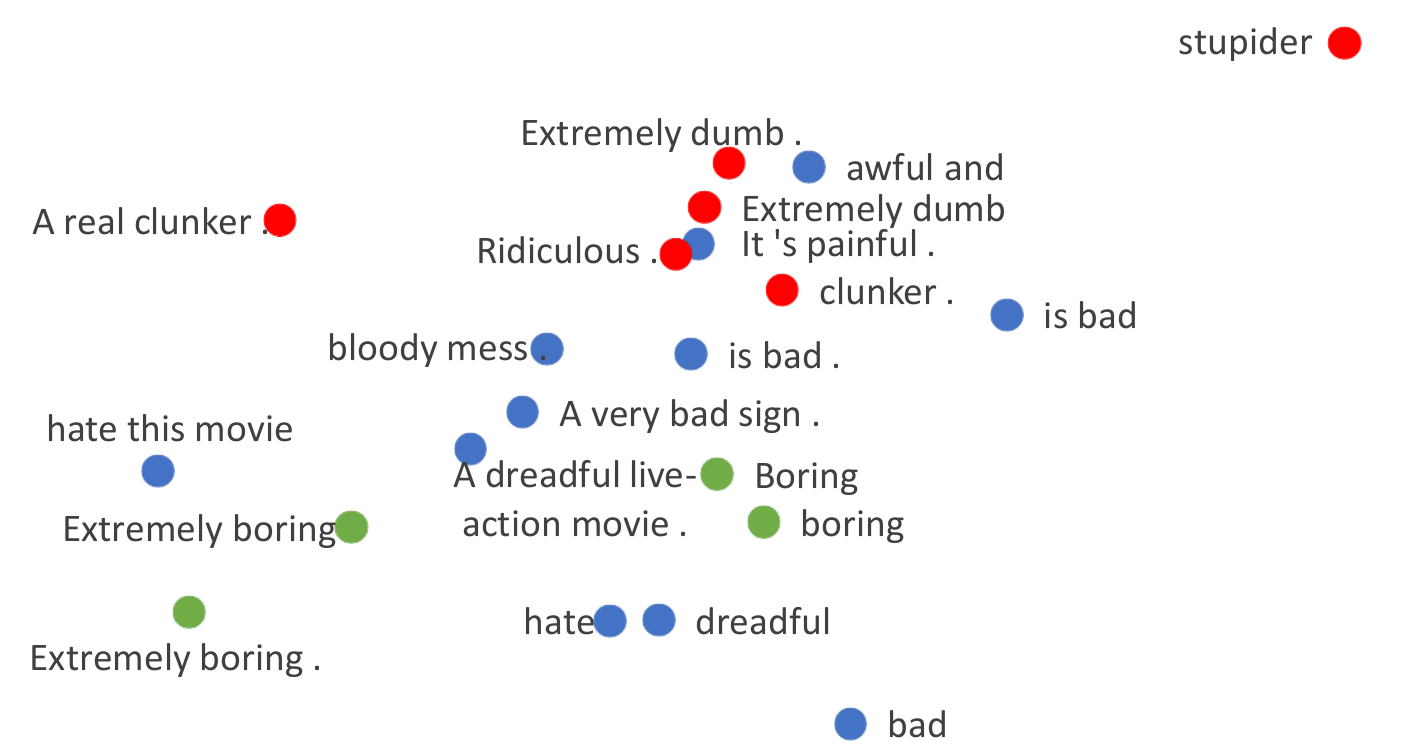}
		\caption{Visualization of correlation of phrases not longer than 5 in the strong negative sentiment space} 
		\label{fig:lveg_pot} 
	\end{minipage} 
\end{figure}

\paragraph{Visualization of sentiment subtypes}
To investigate whether our LVeG can accurately model different emotional subtypes, we visualize all the strong negative sentiment phrases with length below 6 that are classified correctly in a 2D space.
Since in LVeG, 2-dimension 1-component Gaussian mixtures are used to model a distribution over subtypes of a specific sentiment of phrases, we directly represent phrases by their Gaussian means $\bmu$.
From Figure~\ref{fig:lveg_pot}, we see that boring emotions such as ``Extremely boring'' and ``boring''  (green dots) are located at the bottom left, 
stupid emotions such as ``stupider'' and ``Ridiculous'' (red dots) are mainly located at the top right and negative emotions with no special emotional tendency such as ``hate'' and ``bad'' (blue dots) are evenly distributed throughout the space.
This demonstrates that LVeG can capture sentiment subtypes.

\section{Conclusion}
We presented a range of sentiment grammars for using neural networks to model sentiment composition explicitly, 
and empirically showed that explicit modeling of sentiment composition with fine-grained sentiment subtypes gives better performance compared to state-of-the-art neural network models in sentiment analysis.
By using EMLo embeddings, our final model improves fine-grained accuracies by 1.3 points compare to the current best result.

\section*{Acknowledgments}
	This work was supported by the Major Program of Science and Technology Commission Shanghai Municipal (17JC1404102) and NSFC (No. 61572245) . 
    We would like to thank the anonymous reviewers for their careful reading and useful comments.
\bibliography{acl2019}

\begin{thebibliography}{24}
\expandafter\ifx\csname natexlab\endcsname\relax\def\natexlab#1{#1}\fi

\bibitem[{Choi and Cardie(2008)}]{choi2008learning}
Yejin Choi and Claire Cardie. 2008.
\newblock Learning with compositional semantics as structural inference for
  subsentential sentiment analysis.
\newblock In \emph{Proceedings of the conference on Empirical Methods in
  Natural Language Processing}, pages 793--801. Association for Computational
  Linguistics.

\bibitem[{Dong et~al.(2015)Dong, Wei, Liu, Zhou, and Xu}]{dong2015statistical}
Li~Dong, Furu Wei, Shujie Liu, Ming Zhou, and Ke~Xu. 2015.
\newblock A statistical parsing framework for sentiment classification.
\newblock \emph{Computational Linguistics}, pages 265--308.

\bibitem[{Gupta and Zhang(2018)}]{gupta2018attend}
Amulya Gupta and Zhu Zhang. 2018.
\newblock To attend or not to attend: A case study on syntactic structures for
  semantic relatedness.
\newblock In \emph{Proceedings of the 56th Annual Meeting of the Association
  for Computational Linguistics}, pages 2116--2125. Association for
  Computational Linguistics.

\bibitem[{Johnson(1998)}]{johnson1998pcfg}
Mark Johnson. 1998.
\newblock Pcfg models of linguistic tree representations.
\newblock \emph{Computational Linguistics}, pages 613--632.

\bibitem[{Kingma and Ba(2014)}]{kingma2014adam}
Diederik Kingma and Jimmy Ba. 2014.
\newblock Adam: A method for stochastic optimization.
\newblock \emph{International Conference on Learning Representations}.

\bibitem[{Le and Zuidema(2015)}]{le2015compositional}
Phong Le and Willem Zuidema. 2015.
\newblock Compositional distributional semantics with long short term memory.
\newblock In \emph{Proceedings of the Fourth Joint Conference on Lexical and
  Computational Semantics}, pages 10--19. Association for Computational
  Linguistics.

\bibitem[{Li et~al.(2015)Li, Luong, Jurafsky, and Hovy}]{li2015tree}
Jiwei Li, Thang Luong, Dan Jurafsky, and Eduard Hovy. 2015.
\newblock When are tree structures necessary for deep learning of
  representations?
\newblock In \emph{Proceedings of the 2015 Conference on Empirical Methods in
  Natural Language Processing}, pages 2304--2314. Association for Computational
  Linguistics.

\bibitem[{Matsuzaki et~al.(2005)Matsuzaki, Miyao, and
  Tsujii}]{matsuzaki2005probabilistic}
Takuya Matsuzaki, Yusuke Miyao, and Jun{'}ichi Tsujii. 2005.
\newblock Probabilistic {CFG} with latent annotations.
\newblock In \emph{Proceedings of the 43rd Annual Meeting of the Association
  for Computational Linguistics}, pages 75--82, Ann Arbor, Michigan.
  Association for Computational Linguistics.

\bibitem[{McCann et~al.(2017)McCann, Bradbury, Xiong, and
  Socher}]{mccann2017learned}
Bryan McCann, James Bradbury, Caiming Xiong, and Richard Socher. 2017.
\newblock Learned in translation: Contextualized word vectors.
\newblock In \emph{Advances in Neural Information Processing Systems}, pages
  6294--6305.

\bibitem[{Mehrabian(1980)}]{mehrabian1980basic}
Albert Mehrabian. 1980.
\newblock \emph{Basic dimensions for a general psychological theory:
  Implications for personality, social, environmental, and developmental
  studies}.
\newblock Oelgeschlager, Gunn \& Hain Cambridge, MA.

\bibitem[{Moilanen and Pulman(2007)}]{moilanen2007sentiment}
Karo Moilanen and Stephen Pulman. 2007.
\newblock Sentiment composition.
\newblock In \emph{Proceedings of Recent Advances in Natural Language
  Processing}, pages 378--382.

\bibitem[{Montague(1974)}]{Montague1974-MONFPS}
Richard Montague. 1974.
\newblock \emph{Formal Philosophy; Selected Papers of Richard Montague}.
\newblock New Haven: Yale University Press.

\bibitem[{Ortony and Turner(1990)}]{Ortony1990-ORTWBA}
Andrew Ortony and Terence~J. Turner. 1990.
\newblock What's basic about basic emotions?
\newblock \emph{Psychological Review}, 97(3):315--331.

\bibitem[{Pennington et~al.(2014)Pennington, Socher, and
  Manning}]{pennington2014glove}
Jeffrey Pennington, Richard Socher, and Christopher Manning. 2014.
\newblock Glove: Global vectors for word representation.
\newblock In \emph{Proceedings of the 2014 conference on Empirical Methods in
  Natural Language Processing}, pages 1532--1543. Association for Computational
  Linguistics.

\bibitem[{Peters et~al.(2018)Peters, Neumann, Iyyer, Gardner, Clark, Lee, and
  Zettlemoyer}]{peters2018deep}
Matthew Peters, Mark Neumann, Mohit Iyyer, Matt Gardner, Christopher Clark,
  Kenton Lee, and Luke Zettlemoyer. 2018.
\newblock Deep contextualized word representations.
\newblock In \emph{Proceedings of the 2018 Conference of the North American
  Chapter of the Association for Computational Linguistics: Human Language
  Technologies}, pages 2227--2237. Association for Computational Linguistics.

\bibitem[{Petrov et~al.(2006)Petrov, Barrett, Thibaux, and
  Klein}]{petrov2006learning}
Slav Petrov, Leon Barrett, Romain Thibaux, and Dan Klein. 2006.
\newblock Learning accurate, compact, and interpretable tree annotation.
\newblock In \emph{Proceedings of the 21st International Conference on
  Computational Linguistics and the 44th annual meeting of the Association for
  Computational Linguistics}, pages 433--440. Association for Computational
  Linguistics.

\bibitem[{Petrov and Klein(2007)}]{petrov2007improved}
Slav Petrov and Dan Klein. 2007.
\newblock Improved inference for unlexicalized parsing.
\newblock In \emph{Human Language Technologies 2007: The Conference of the
  North American Chapter of the Association for Computational Linguistics;
  Proceedings of the Main Conference}, pages 404--411. Association for
  Computational Linguistics.

\bibitem[{Socher et~al.(2013)Socher, Perelygin, Wu, Chuang, Manning, Ng, and
  Potts}]{socher2013recursive}
Richard Socher, Alex Perelygin, Jean Wu, Jason Chuang, Christopher~D Manning,
  Andrew Ng, and Christopher Potts. 2013.
\newblock Recursive deep models for semantic compositionality over a sentiment
  treebank.
\newblock In \emph{Proceedings of the 2013 conference on Empirical Methods in
  Natural Language Processing}, pages 1631--1642. Association for Computational
  Linguistics.

\bibitem[{Taboada et~al.(2011)Taboada, Brooke, Tofiloski, Voll, and
  Stede}]{taboada2011lexicon}
Maite Taboada, Julian Brooke, Milan Tofiloski, Kimberly Voll, and Manfred
  Stede. 2011.
\newblock Lexicon-based methods for sentiment analysis.
\newblock \emph{Computational linguistics}, pages 267--307.

\bibitem[{Tai et~al.(2015)Tai, Socher, and Manning}]{tai2015improved}
Kai~Sheng Tai, Richard Socher, and Christopher~D. Manning. 2015.
\newblock Improved semantic representations from tree-structured long
  short-term memory networks.
\newblock In \emph{Proceedings of the 53rd Annual Meeting of the Association
  for Computational Linguistics and the 7th International Joint Conference on
  Natural Language Processing}, pages 1556--1566. Association for Computational
  Linguistics.

\bibitem[{Teng et~al.(2016)Teng, Vo, and Zhang}]{teng2016context}
Zhiyang Teng, Duy~Tin Vo, and Yue Zhang. 2016.
\newblock Context-sensitive lexicon features for neural sentiment analysis.
\newblock In \emph{Proceedings of the 2016 Conference on Empirical Methods in
  Natural Language Processing}, pages 1629--1638. Association for Computational
  Linguistics.

\bibitem[{Teng and Zhang(2017)}]{teng2017head}
Zhiyang Teng and Yue Zhang. 2017.
\newblock Head-lexicalized bidirectional tree lstms.
\newblock \emph{Transactions of the Association for Computational Linguistics},
  5:163--177.

\bibitem[{Zhao et~al.(2018)Zhao, Zhang, and Tu}]{zhao2018gaussian}
Yanpeng Zhao, Liwen Zhang, and Kewei Tu. 2018.
\newblock Gaussian mixture latent vector grammars.
\newblock In \emph{Proceedings of the 56th Annual Meeting of the Association
  for Computational Linguistics}, pages 1181--1189. Association for
  Computational Linguistics.

\bibitem[{Zhu et~al.(2015)Zhu, Sobihani, and Guo}]{zhu2015long}
Xiaodan Zhu, Parinaz Sobihani, and Hongyu Guo. 2015.
\newblock Long short-term memory over recursive structures.
\newblock In \emph{International Conference on Machine Learning}, pages
  1604--1612.

\end{thebibliography}
\bibliographystyle{acl_natbib}

\end{document}